\documentclass[preprint,3p, twocolumn]{elsarticle}
\usepackage[english]{babel}
\usepackage{microtype}

\usepackage{latexsym}

\usepackage{graphicx}
\usepackage[T1]{fontenc}

\usepackage{amsmath}
\usepackage{amsfonts}
\usepackage{amssymb}
\usepackage{amsbsy}
\usepackage{amsthm}
\usepackage{enumitem}
\usepackage{booktabs}
\usepackage{multirow}
\usepackage{subcaption}
\usepackage{siunitx}
\usepackage{listings}

\usepackage[pdftex,colorlinks=true,urlcolor=blue,citecolor=black,anchorcolor=black,linkcolor=black]{hyperref}
\usepackage[noabbrev,capitalise]{cleveref}

\begin{document}

\begin{frontmatter}

\title{Simulation Code Generation for Fluid Systems using Large Language Models: \\ Benchmarking Models and Prompting Strategies}

\author[1,2]{J. Marius Stürmer\corref{cor1}}
\ead{jan.stuermer@dlr.de}
\author[1]{Jascha Knack}
\author[1]{Tobias Koch}
\author[2]{Andreas Weinmann}
\cortext[cor1]{Corresponding author}

\affiliation[1]{organization={German Aerospace Centre (DLR), Institute for the Protection of Terrestrial Infrastructures},
addressline={Rathausallee 12},
postcode={53757},
city={Sankt Augustin},
country={Germany}}
\affiliation[2]{organization={Algorithms for Computer Vision, Imaging and Data Analysis, Technische Hochschule Würzburg-Schweinfurt},
addressline={Ignaz-Schön-Straße 11},
city={Schweinfurt},
postcode={97421},
country={Germany}}

\begin{abstract}
Large language models (LLMs) have demonstrated a strong ability to generate syntactically correct code from natural-language specifications. In this study, we explore how LLMs can be harnessed to automatically translate a neutral graph representation of fluid‑system models into executable code for two widely adopted simulation environments: the Python library \texttt{WNTR} and the Modelica Standard Library.
We conduct a systematic comparison of ten state‑of‑the‑art LLMs and six prompting strategies that differ in the contextual information supplied (e.g., code or documentation). For each configuration we assess the generated code using a suite of software‑quality metrics and we validate the functional fidelity of the resulting simulation models by reproducing benchmark fluid‑system scenarios.
Our findings offer concrete guidance for researchers and engineers seeking to integrate LLM‑driven code synthesis into model‑based design pipelines. While the best‑performing configurations achieve acceptable syntactic quality, we observe substantial gaps remain in simulation fidelity.
\end{abstract}


\end{frontmatter}

\section{Introduction}

Engineering disciplines that deal with hydraulics and fluid-system design depend heavily on simulations.
As highlighted by \citet{Banks.2010}, simulation offers a powerful approach to modeling complex systems, enabling experimentation with policies and designs and gaining insight into the system behavior.
The challenge of effectively integrating domain knowledge into such systems has long been recognized \cite{Novak.2015}.
In domains such as fluid systems and energy infrastructure, simulations are indispensable for enabling cost-effective prototyping and robust analysis under real-world conditions \cite{Makahleh.2024}.
In combination with Digital Twins, simulations can even be leveraged to raise critical infrastructure resilience \cite{Lampropoulos.2024} in, e.g., residential water demand \cite{Sattler.2023}.

Constructing these simulation models, however, remains a manual, expertise-driven activity. 
The implementation of simulation models is time-consuming and requires domain expertise \cite{Cellier.2006}, while it is also prone to human error \cite{Oberkampf.2002} and must often be repeated for each new system configuration or variant.
Recent studies have shown that engineering schematics, particularly Piping and Instrumentation Diagrams (P\&IDs), can be automatically transformed into neutral graph representations \cite{Stuermer.2025}.
This capability opens the door to a range of downstream automation opportunities, most notably the generation of simulation models directly from the extracted graph.
Early approaches to automated simulation generation relied on template-based code generation, such as translating SysML specifications into Simulink models \cite{Pohlmann.2011} or creating Modelica-based building energy simulations \cite{NytschGeusen.2017}. 
Other examples of automated simulation generation from graphs include methods proposed by \citet{Martinez.2018} and \citet{Stuermer.2023}.
Despite these advances, a bottleneck remains: each engineering-simulation framework (e.g., Modelica, WNTR, Simulink) defines its own set of domain-specific abstractions, modeling primitives, and code conventions. 
Consequently, existing automated pipelines are usually bound to a single environment and depend on handcrafted translation rules. 
This limits generalizability and slows the integration of new simulation tools or domain libraries.

Large Language Models (LLMs) offer a promising alternative.
Recent applications of techniques like RAG in domains such as aviation model design \cite{Xiong.2025} demonstrate the potential of AI-driven approaches. 
LLMs show strong performance in generating code from natural language and understanding domain knowledge \cite{Chen.2021}, and thus have the potential to replace rule-based translators with a more flexible approach.
Recent work by \citet{Jin.2026} highlights that LLMs often struggle to reliably align generated code with natural language specifications, frequently misclassifying correct implementations as non-compliant. 
This underscores the need for careful prompt design and validation mechanisms when applying LLMs to domain-specific tasks. 
Prior work on LLM-generated simulation code has also shown mixed results \cite{Xiang.2025, Ren.2025}.

In this work, we refer to the automatically generated, framework-specific code modules as simulation interface layers containing multiple wrapper functions, one for each component.
These layers translate a neutral, graph-based representation of a fluid system into executable code for a particular simulation environment.

\begin{figure*}[tb]
    \centering
    \includegraphics[width=0.92\linewidth]{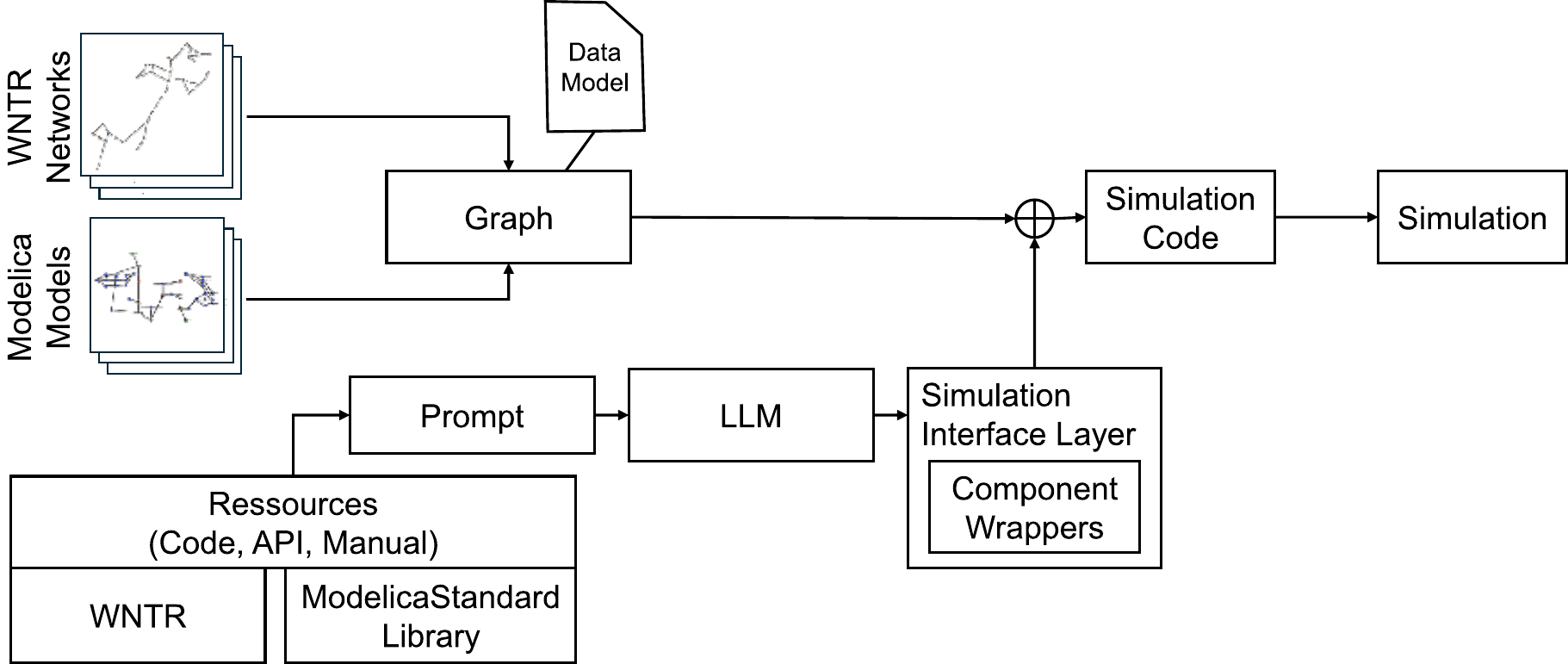}
    \caption{Pipeline for automatic generation of simulation-interface layers. A neutral graph of a hydraulic system is combined with contextual material (API docs, stubs, or curated code) and fed to an LLM, which synthesizes a framework-specific wrapper (WNTR Python module or Modelica file).  The generated code is then compiled/executed and validated against the ground-truth model.}
    \label{fig:project_pipeline}
\end{figure*}

\paragraph{Contributions} This paper makes three key contributions:
\begin{enumerate}
  \item \textbf{Systematic evaluation of LLMs for simulation code generation.} We compare state-of-the-art models and prompting strategies across two distinct simulation frameworks (WNTR \cite{Klise.2020} and Modelica \cite{ModelicaAssociation.2025}), revealing domain-specific performance patterns and limitations.

  \item \textbf{Analysis of knowledge integration strategies.} Through controlled experiments with varying contextual information (documentation, full code, stubs, curated examples), we demonstrate that relevance outweighs volume in prompt engineering.

  \item \textbf{Assessment of agent-based refinement.} We evaluate OpenCode with validation feedback, showing that while agentic approaches improve syntactic correctness, they cannot resolve deeper semantic issues affecting simulation fidelity.
\end{enumerate}

The remainder of the paper is organized as follows.
\cref{sec:sota} describes related work on LLM-based code generation and prior attempts at automating simulation model creation.  
\cref{sec:method} details the methodology for knowledge integration, prompt construction, synthesis of simulation interface layers, the benchmark dataset, experimental design, and evaluation metrics.
\cref{sec:results} presents the results across LLMs and knowledge integration strategies.  
Finally, \cref{sec:conclusion} discusses the implications of our findings, outlines current limitations, and suggests directions for future research.

\section{Related Work}
\label{sec:sota}

Large Language Models (LLMs) have recently become a commonly used tool for general-purpose code synthesis  \cite{Jiang.2025}, yet their use for automatically producing simulation-specific code, which demands precision, domain-specific APIs, and adherence to physical constraints remains underexplored.
Accordingly, this section reviews LLM-based code-generation techniques as well as existing methods that translate structured engineering representations into simulation models and for augmenting LLMs with retrieval or graph-based knowledge of existing code bases and APIs \cite{Chen.2026}.

\paragraph{LLMs for Code Generation}
LLMs as the de-facto standard for general-purpose code synthesis are embedded in many coding assistants \cite{Huynh.2025, Bistarelli.2025}. 
While straight forward ``text-to-code'' prompting works well for isolated snippets, it often fails when the generated code must respect a project's type system, naming conventions, or a complex API surface.
This limitation has motivated a line of research on repository-level or project-aware generation.
Such systems first retrieve relevant files, parse abstract syntax trees (ASTs) or call graphs, and then condition the LLM on this multi-file context.
Examples of such systems are RepoCoder \cite{Zhang.2023} or RepoMinCoder \cite{Li.2024b}.
These repository-aware frameworks reduce API misuse and improve generation consistency by explicitly incorporating multi-file context into retrieval and prompt construction.

While effective for their target domains, these approach-es lack the flexibility to generalize across simulation frameworks or arbitrary network topologies.

Retrieval-Augmented Generation (RAG) couples an LLM with an external knowledge store, allowing the model to retrieve relevant context at inference time.
In the software-engineering context, the retrieved artefacts are typically functions, docstrings, or examples drawn from the target repository and injected into the prompt \cite{Gao.2023}.
Recent work augments this idea with explicit graph structures (GraphRAG) which exploit relational information such as call-graphs, dependency graphs, or requirement graphs to retrieve context that is structurally aligned with the generation task \cite{Han.2024}.

Agentic frameworks even extend retrieval-augmented generation by introducing tool use, iterative refinement, and automated testing. 
An LLM (or a hierarchy of LLMs) orchestrates sub-agents that explore a repository, retrieve relevant code, execute tests, and iteratively repair errors, enabling reasoning over entire code bases rather than isolated snippets \cite{Li.2025, Zhang.2024}.
Another popular Open Source Tool that provides these functionalities is OpenCode \cite{anomalyco.2026}.

\paragraph{Automated Graph to Simulation Pipelines}
Long before LLMs, there are approaches to automate the process of simulation model creation by deriving executable models from high-level representations, such as P\&IDs, GIS data, or standardized graph formats (CAEX, DEXPI, EPANET) \cite{Drath.2022, Tolksdorf.2025, Rossman.2012}. 

Classical approaches \cite{Stuermer.2023, Kaiser.2022, Bjrnskov.2025} parse a domain-specific language or graph and apply deterministic, template-based code generators to emit Modelica, Simulink, or Python code.
\citet{Mans.2022} present a framework that builds district-heating and cooling network models from geographic and graph data and automatically exports them as Modelica models for downstream simulation. 
\citet{Bahamdan.2026} use multi-agent LLMs to convert process diagrams into Aspen HYSYSs, thus handling the diagram to graph and then graph to simulation process implicitly.

For fluid systems, several standardized graph-based representations exist, such as CAEX (AutomationML) \cite{Drath.2022}, \texttt{DEXPI} \cite{Tolksdorf.2025}, or the input formats for water networks like EPANET \cite{Rossman.2012}.
Furthermore, modeling languages like SysML and ModelicaML have been proposed to bridge the gap between system requirements and executable simulation models \cite{Pohlmann.2011}. 

\paragraph{LLM-Based Generation of Simulation Code}
A growing number of recent work targets the direct generation of simulation code (Modelica, Simulink, or domain-specific Python libraries) from textual or diagrammatic specifications.  

SimuGen translates engineering diagrams into Simulink models by chaining specialized agents (parser, generator, tester, executor) and a block-library database, achieving high fidelity on a curated Simulink benchmark \cite{Ren.2025}.
The authors report high reproduction accuracy on their Simulink dataset, demonstrating the benefit of multimodal input and agentic verification.
Another work by \citet{Xia.2024b} similarly uses multi-agent systems for parametrization of simulations.

Text2Model fine-tunes an LLM to produce dynamic chemical-reactor models in Modelica from natural-language descriptions; the study reports frequent unit-conversion and structural-equation errors, but demonstrates that domain-specific fine-tuning of an small LLM markedly improves compilation and simulation success \cite{Rupprecht.2025}.

ModiGen combines supervised fine-tuning, GraphRAG, and reinforcement-learning-based feedback to boost Modelica generation quality, again highlighting the importance of external knowledge integration \cite{Xiang.2025}.
Similarly, recent work on computational fluid dynamics demonstrates the potential of domain-adapted LLMs by fine-tuning a small LLM and highlight the critical role of domain-specific adaptation to handle complex engineering workflows \cite{Dong.2025}.

Collectively, these papers indicate that naïve prompting of general-purpose LLMs is insufficient for high-quality simulation code.
As will be explored in this work, the challenge is not just code generation, but bridging the gap between high-level system descriptions and executable, physically accurate simulations.

\paragraph{Positioning our Work}
Existing LLM based research on simulation code generation typically focuses on creating self-contained Modelica components from textual descriptions, rather than developing adapter modules that are required to integrate with large existing codebases while following established API patterns. 
Our work addresses this gap by systematically comparing different knowledge integration strategies, particularly examining how documentation can provide essential context for accurate code generation.

While general code generation has been extensively studied, simulation code presents unique challenges where small implementation details critically impact functionality. 
Our study contributes to this underexplored area by evaluating both smaller models and large models across multiple prompting strategies, providing insights into the current capabilities and limitations of LLM-based simulation code synthesis.

\section{Methodology}
\label{sec:method}

\begin{figure*}[tb]
  \centering
  \includegraphics[width=0.92\linewidth]{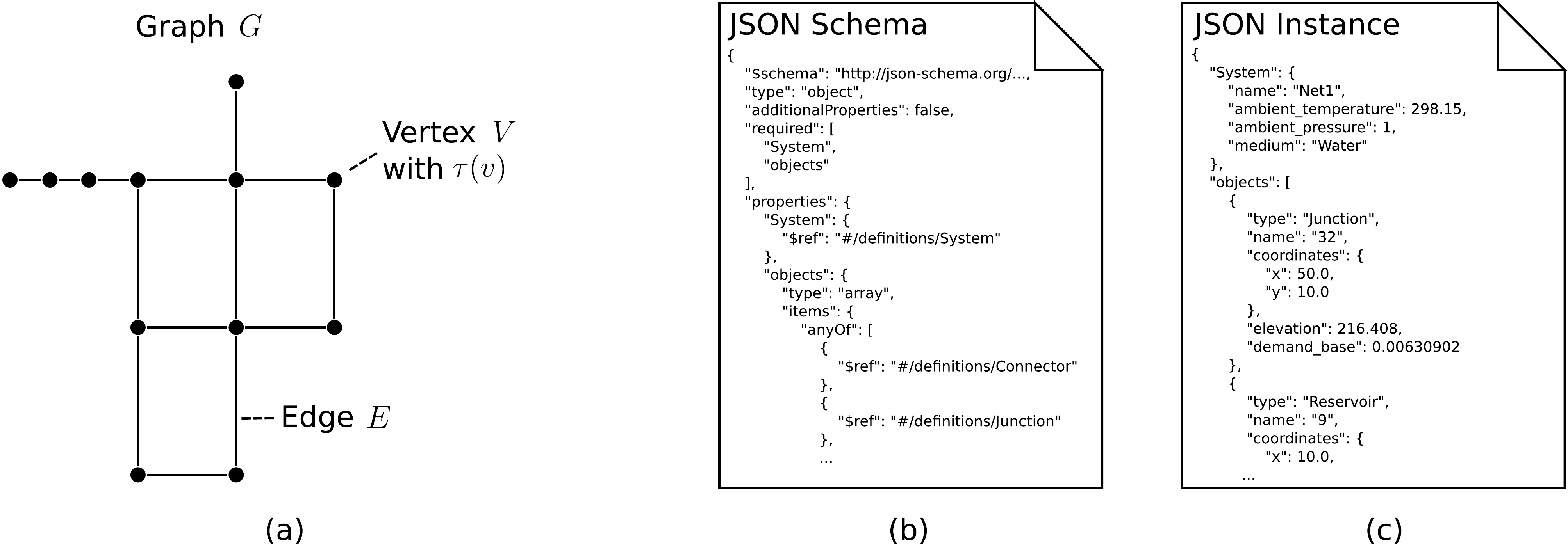}
  \caption{(a) Example WNTR water network topology illustrating basic graph terminology. (b) JSON schema snippet. (c) JSON instance snippet corresponding to the network in (a).}
  \label{fig:two_networks}
\end{figure*}

The objective of this study is to generate a ``simulation-interface layer'' that converts a neutral, graph-based description of a fluid system into executable code for two widely used simulation environments, the Python library \texttt{WNTR}, which focuses on simulating water networks, and the Modelica Standard Library (MSL), which is mostly used for modeling fluid systems.
Both environments expose an API that must be used by the generated wrapper to generate correct code.
The paper concept is displayed in \cref{fig:project_pipeline} and each part will be described in the following steps.

\subsection{Problem Definition}
We represent a fluid system as a labeled, directed multigraph $G = (V, E, \tau)$ (see \cref{fig:two_networks}), where
\begin{itemize}
  \item $V$ is the set of component vertices (e.g., pumps, pipes, valves, tanks);
  \item $E \subseteq V \times V$ encodes the directed flow connections between components;
  \item $\tau : V \cup E \rightarrow \mathcal{A}$ is a labeling function that assigns to each vertex or edge a dictionary of parameters $\mathcal{A}$ including a type tag.
\end{itemize}
As an illustration, a pump vertex $v_{\text{pump}}$ may be annotated as  
\begin{align*}
\tau(v_{\text{pump}})=\bigl\{&\texttt{type}= \texttt{Pump}, \\
  &\texttt{efficiency}=0.85, \\
  &\texttt{speed}=1500~\text{rpm} \\
  &... \bigr\}
\end{align*}
For each target simulation environment we posit an a priori unknown mapping wrapper from the graph to simulation-environment specific code, either

\begin{equation*}
\mathcal{W}: G \longrightarrow \texttt{WNTR\_code,} \qquad 
\end{equation*}
or
\begin{equation*}
\mathcal{M}: G \longrightarrow \texttt{Modelica\_code} \text{.}
\end{equation*}

In the first case, $\mathcal{W}$ is required to emit a Python WNTR network object after (i)~importing WNTR, (ii)~instantiating the appropriate network objects with connections according to the edges $E$, and (iii)~assigning all required parameters from $\tau$, performing appropriate unit conversions.
In the second case, $\mathcal{M}$ is required to produce a Modelica text file that (i)~imports the relevant MSL packages, (ii)~declares component instances, (iii)~connects them using \texttt{connect} equations according to the edges $E$, and (iv)~sets the corresponding parameter values from $\tau$, performing appropriate unit conversions.

A generated wrapper $\mathcal{W}$ or $\mathcal{M}$ is deemed acceptable if (i) the code parses without errors, (ii) every call to a framework respects the signature (names, ordering, and expected units) and (iii) when the generated model is simulated under a predefined set of inlet/outlet boundary conditions, the observed timeseries (pressures, flows, heads) equal those of a reference ground-truth up to a tolerance of $\varepsilon$.

\subsection{Graph Specification}
\label{sec:specification}
To enable a framework-agnostic creation of code, we require a neutral representation of the systems that serves as the data model for the networks and as input for the LLM.
It is neutral in the sense that it is required to describe the topology of the network, the specific component types, and their associated physical parameters, while remaining independent of the target simulators.
In \cref{fig:project_pipeline}, this refers to the Graph and its Data Model and in \cref{fig:two_networks}, the JSON schema and the JSON instance.

As described in \cref{sec:sota}, several powerful standards to describe such graphs exist, but they are often tightly coupled to specific tool chains or contain extensive metadata unnecessary for the specific task of code generation.
For this study, we designed a custom JSON-based schema that prioritizes the intersection of requirements for our two target environments: WNTR and MSL. 
The schema is constructed based on the following principles: 
\begin{enumerate} 
    \item \textbf{Intersectionality:} Only parameters required by both target frameworks are mandatory fields. Parameters specific to one framework (e.g., specific initialization options) are treated as optional or omitted. 
    \item \textbf{Explicit Typing:} Every node carries an explicit class tag (e.g., \textit{Pump}, \textit{Valve}, \textit{Junction}).
    \item \textbf{Unit Awareness:} Numerical values are accompanied by unit definitions (e.g., \textit{``unit'': ``m''}). 
    \item \textbf{Structural Clarity:} The graph is serialized as a list of objects (nodes) with implicit connectivity defined via named references.
\end{enumerate}

The serialization follows a JSON structure containing a \textit{System} object (defining global ambient conditions) and an array of \textit{Objects} defining the components.
The schema supports a wide variety of components, including standard elements (such as Pipes, Junctions, Reservoirs) as well as control logic elements (such as Sensors, Hysteresis blocks, Gain blocks).

% \cref{lst:schema} provides the complete JSON Schema definition used in our benchmark suite. 

\subsection{Context \& Prompts}
\label{sec:prompts}

While the graph representation provides a framework-agnostic input format, the quality of generated simulation code depends on how domain knowledge is encoded in the prompt. 
We evaluate different prompting configurations that progressively increase the contextual information available to the LLM.
All prompts share a common core that conveys the task and the data model to the language model existing of a task description, a schema block, an example network block, an instruction and a context block. 
An example of the prompt can be seen in \cref{fig:prompt}.

\begin{figure}[bt]
\centering
\begin{subfigure}[t]{0.45\textwidth}
\begin{lstlisting}[label={lst:prompt_wntr}, frame=single]
I am an engineer and I want to create simulation code easily with the python library 'WNTR'.
I need you to write code that transforms a graph structured system into python code to have a runnable WNTR simulation.
It should not depend on the specific example but provide generic wrapper functionality.
Each component type mapped from the data_model to WNTR should have a specific "wrapper" function called "wrap_xxx(wn, obj:dict, start:str (optional), end:str (optional))".
Do not write wrapper functions as class functions.
No blabbering.
Data model schema: ...
Example Network: ...
Context: ...
\end{lstlisting}
\end{subfigure}
\caption{The instruction asks the model to produce generic functions that translate a graph-structured system description (provided via a JSON schema and an example network) into executable Python code using the WNTR library. The prompt explicitly forbids class-based wrappers and requests a reusable implementation. \label{fig:prompt}}
\end{figure}

For the context block, we explore five configurations of increasing complexity.
This results in five different prompt configurations.
Note that for MSL, we only include files that are either relevant for math, fluids or control blocks.
If multiple files are used, all are concatenated in a single block, preceded by their file names.
To optimize token usage, we compress the files by removing redundant whitespace and excluding non-essential elements such as test functions.
The prompt is then sent to the LLM in a single request.

\paragraph{Prompt Configuration 1, Zero-Shot (no context)}  
Only the core elements are provided, no additional documentation, stubs, or code examples are included.

\paragraph{Prompt Configuration 2, Documentation} 
For WNTR, we use a set of RST-files from the its repository. For Modelica, the HTML-annotated class signatures from the MSL are used.

\paragraph{Prompt Configuration 3, Stubs}  
A minimal, syntactically valid skeleton of the target wrapper is provided, containing only function signatures and return type declarations. 
All implementation logic is omitted.
For Modelica, e.g., connection statements and equation blocks are specifically excluded.

\paragraph{Prompt Configuration 4, Curated / selected code}  
To isolate the most informative pieces of the library, we manually extract a compact subset of the source that include the necessary API calls. 
This curated snippet is the smallest amount of code that still conveys the logic required for correct wrapper generation.

\paragraph{Prompt Configuration 5, Full Code}  
We attach the complete set of source files that constitute the target library.
For WNTR these are all non-trivial Python files and for Modelica, every \texttt{.mo} file required for the fluid, math, and logical components is included.

\paragraph{Coding Agent}
\label{sec:opencode_experiment}

As described in \cref{sec:sota}, earlier work (e.g., \cite{Bahamdan.2026, Xia.2024b}) has successfully used coding agents on related tasks. 
To mimic these multi-agent approaches, we implement OpenCode \cite{anomalyco.2026} with access to the WNTR source code and the Modelica Standard Library source.
The setup includes running the experiment in two variants.
First, we run OpenCode with the standard prompt (prompt configuration 1).
Secondly, to use the self-validation capabilities, OpenCode was instructed to iterate until a validation loop passes successfully.
We define a validation function that can be passed only if the simulation interface layer generates usable output.
The validation function only checks if the code runs, and not wether it is correct or not. 
This way, the coding agent can re-iterate until the code runs.

\begin{table*}[tb]
\centering
\caption{Size of the contextual material supplied to the models for each prompting strategy.}
\begin{tabular}{@{}lrrrr@{}}
\toprule
\multirow{2}{*}{Context} & \multicolumn{2}{c}{WNTR}     & \multicolumn{2}{c}{Modelica} \\
                         & Word Count & Character Count & Word Count & Character Count \\ \midrule
Full Code                & \num{84829} & \num{948230} & \num{55973} & \num{755303} \\
Stubs                    & \num{36668} & \num{320016} & \num{50142} & \num{562306} \\
Documentation            & \num{32195} & \num{275170} & \num{86542} & \num{618762} \\
Curated Code             & \num{6242} & \num{60256} & \num{5477} & \num{78752} \\
None (Zero-Shot)         & -& -& -& -\\ \bottomrule
\end{tabular}
\end{table*}

\begin{figure*}[tb]
    \centering
    \includegraphics[width=0.8\linewidth]{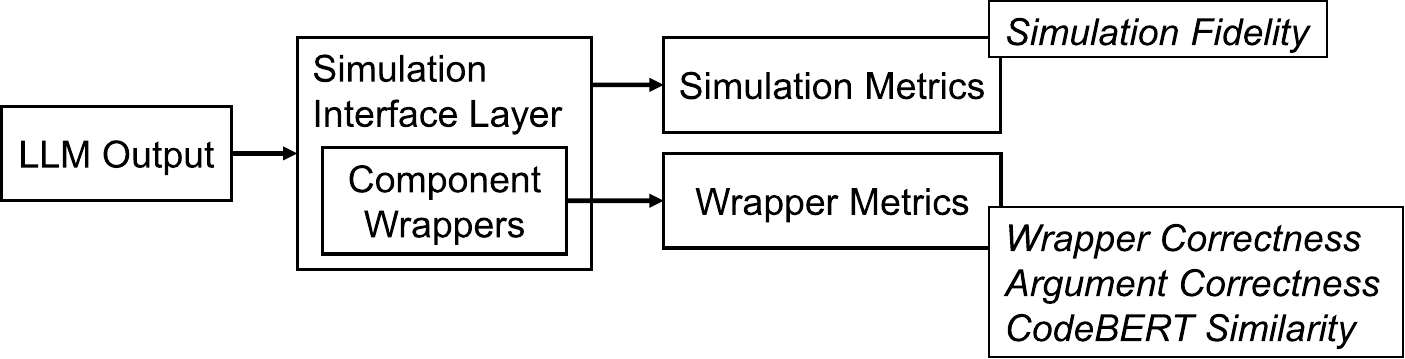}
    \caption{Overview of the evaluation pipeline. The generated wrapper is first parsed (syntactic check), then token-wise compared to the ground-truth (argument check), and finally executed. Its output trajectories are compared to the reference simulation (semantic check).}
    \label{fig:eval_pipeline} % TODO: Call Correctness hinzufügen und als PDF einfügen
\end{figure*}

\subsection{Evaluation Metrics}
\label{sec:metrics}

Assessing the quality of LLM-generated simulation code requires an approach that captures both syntactic correctness and functional fidelity. 
Prior work on code generation has leveraged code clone detection \cite{Svajlenko.2020, MartinezGil.2024} to assess the similarity between generated and ground-truth implementations, typically relying on static syntactic or semantic comparisons.
However, simulation code introduces unique challenges: small errors in parameter mappings or API misuse can lead to semantically invalid models that compile but produce incorrect simulations.
To address this, we combine static analysis (syntactic validity, API conformity) with dynamic validation (simulation fidelity), drawing inspiration from recent LLM-based code generation studies \cite{Xiang.2025, Moltner.2026}.
Our evaluation pipeline, depicted in \cref{fig:eval_pipeline}, integrates these metrics to systematically validate both the syntactic structure and functional behavior of the generated wrappers.
All wrapper-level metrics are computed per-wrapper and then averaged over the $N$ wrappers in the benchmark.

\subsubsection{Wrapper Correctness}
\label{sec:wrapper-correct}

The most fundamental requirement for a generated wrapper is executability: it must import and run without raising exceptions. 
To quantify this, we adopt the pass@k metric from code generation literature \cite{Chen.2021}, which measures the probability that at least one of $k$ generated solutions is executable. 
Given computational constraints imposed by the large number of model configurations and benchmark cases, we limit our evaluation to $3$ runs per configuration.
The pass@1 metric captures the baseline success rate of single-generation attempts and equals the average over the three runs, while pass@3 reflects the best-case performance when selecting from the three runs.

\begin{align*}
\text{pass@1} &= \frac{1}{N}\sum_{i=1}^N E_i \\
\text{pass@3} &= \frac{1}{N}\sum_{i=1}^N \text{max}(E_{i,1}, E_{i,2}, E_{i,3})
\end{align*}

where $E_i$ is defined as $1$ if wrapper $i$ executes successfully with a dummy input and $0$ otherwise. 
Analogously, we define $E_{i,1}$, $E_{i,2}$ and $E_{i,3}$, where the second index corresponds to the respective run.

\subsubsection{Argument Correctness}
\label{sec:arg-correct}

We evaluate the semantic correctness of generated wrappers through two complementary metrics: call-level F1 score and argument-level F1 score. 
The F1 score is the harmonic mean between precision and recall and is calculated using true positives (TP), false positives (FP), and false negatives (FN):
\begin{align*}
\text{F1} &= 2 \times \frac{\text{Precision} \times \text{Recall}}{\text{Precision} + \text{Recall}}
\end{align*}
with $\text{Precision} = \frac{TP}{TP + FP}$ and $\text{Recall} = \frac{TP}{TP + FN}$.
For each wrapper we extract the set of ground-truth calls and generated calls.
Each call is represented as a dictionary containing the function name and its arguments. 
Both call-level F1 and argument-level F1 are computed at both micro (per-wrapper) and macro (global) levels to provide information into model performance.
For macro F1 scores, we compute the F1 score for each wrapper individually and then average across all wrappers. 
For micro F1 scores, we aggregate all true positives, false positives, and false negatives across all wrappers before computing the F1 score.

\begin{figure*}[tb]
    \centering
    \includegraphics[width=0.92\linewidth]{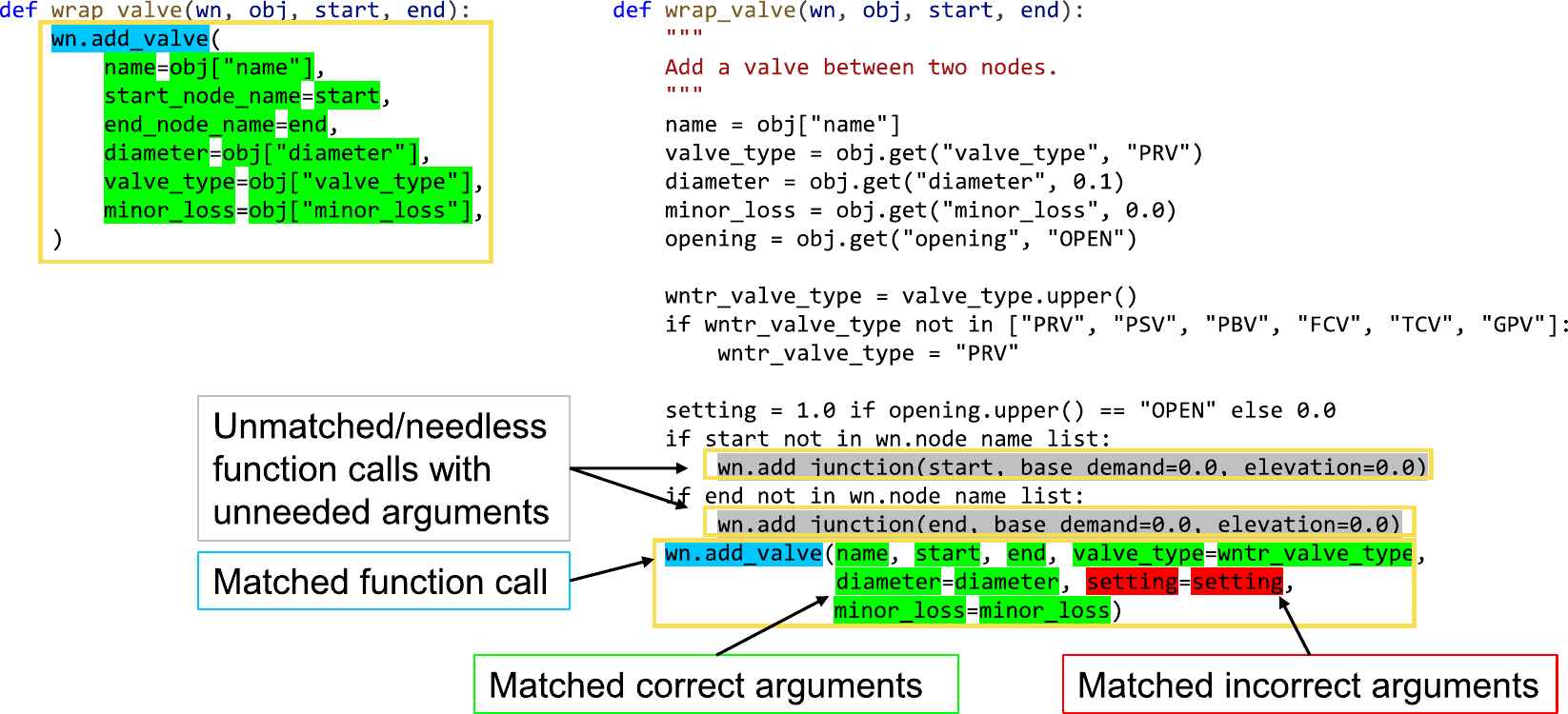}
    \caption{Token-level comparison between ground-truth (left) and LLM-generated (right) functions. Correct calls and arguments are highlighted in green, while missing or mismatched elements appear in red.}
    \label{fig:arg_diff}
\end{figure*}

\begin{itemize}
    \item \textbf{Call-level}: A true positive occurs when a function call appears in both ground truth and generated code;
    \item \textbf{Argument-level}: A true positive occurs when both the argument name and value match between ground truth and generated code
\end{itemize}
Because macro uses an average over all scores, we refer to macro F1 scores as Avg. F1 and to micro F1 scores simply as F1 scores.
\Cref{fig:arg_diff} illustrates a visual comparison between ground-truth and generated functions, highlighting correct calls/arguments in green and mismatches in red.

%%%%%%%%%%%%%%%%% THIS FIGURE BELONGS FURTHER UP 

\subsubsection{Simulation Fidelity}
\label{sec:sim-correct}

Validation of the recreated simulation models is performed by simulating both the reference and the generated model under identical boundary conditions.
For network $k$ ($k=1,\dots,M$) and component $i$ ($i=1,\dots,n_k$) we denote the reference and predicted time series by $y_{k,i}(t)$ and $\hat{y}_{k,i}(t)$, respectively, with $t=1,\dots,T$.
The Normalized Mean Absolute Error (NMAE) for a component is
\begin{align*}
\text{NMAE}_{k,i} &= \frac{\frac{1}{T} \sum_{t=1}^T{\big| y_{k,i}(t) - \hat{y}_{k,i}(t) \big|}}{\max_t y_{k,i}(t) - \min_t y_{k,i}(t)}
\end{align*}
A component is counted as correct if $\text{NMAE}_{k,i}\leq\varepsilon$ with $\varepsilon=0.05$ (5\;\%).
The per-network correctness is given by
\begin{align*}
C_k &= \frac{\sum_{i=1}^{n_k} c_{k,i}}{n_k} \\
\end{align*}
where $c_{k,i} = 1$ if $\text{NMAE}_{k,i} \le 0.05$, and $c_{k,i} = 0$ else.

\subsubsection{CodeBERT Similarity}

To complement the static argument comparison, we measure the semantic similarity between generated and ground-truth wrapper implementations using the pretrained transformer model CodeBERT \cite{Feng.2020}, capable of producing embeddings of code snippets.
This follows the idea of \cite{MartinezGil.2024} where CodeBERT is used for clone code detection.

Given a wrapper implementation as source code string $c$, the CodeBERT encoder $f(\cdot)$ maps the code into a continuous embedding vector $\mathbf{z} = f(c) \in \mathbb{R}^d$ where $d$ is the embedding dimension of the model, in our case $d=768$. 

For each wrapper $i$, we compute embeddings for the ground-truth implementation $c_i^{gt}$ and the generated implementation $c_i^{gen}$:

\begin{equation*}
\mathbf{z}_i^{gt} = f(c_i^{gt}), \qquad
\mathbf{z}_i^{gen} = f(c_i^{gen})
\end{equation*}

The similarity between both implementations is measured using cosine similarity:

\begin{equation*}
S_i =
\frac{\mathbf{z}_i^{gt} \cdot \mathbf{z}_i^{gen}}
{\|\mathbf{z}_i^{gt}\| \, \|\mathbf{z}_i^{gen}\|}
\end{equation*}

Higher values indicate greater semantic similarity between the generated wrapper and the reference implementation.
For a set of $N$ evaluated wrappers, the overall similarity score is computed as the average cosine similarity:
\begin{equation*}
S_{\text{avg}} = \frac{1}{N} \sum_{i=1}^{N} S_i
\end{equation*}
This metric provides a semantic comparison of wrapper implementations that is robust to superficial differences such as formatting, variable names, or minor structural variations.

\begin{table*}[tb]
\centering
\caption{Large-language models evaluated in this study.
The models were chosen for their recency and relevance to code generation. Listed are quantization format, parameter count and maximum context window. Commercial APIs (ChatGPT-5, Gemini-3, Claude) do not expose quantization or context-window information. Model names link to the exact Hugging Face checkpoints used. \label{tab:models}}
\small
\begin{tabular}{@{}llcccc@{}}
\toprule
Model                         & Quantization & BPW & Parameters & Activated Parameters & Context Window \\ \midrule
\href{https://huggingface.co/unsloth/Qwen3-Coder-Next-GGUF}{\texttt{Qwen3-Coder-Next}} & UD-Q8\_K\_XL & 8.61 & 80B     & 3B & 256k \\
\href{https://huggingface.co/unsloth/gpt-oss-120b-GGUF}{\texttt{gpt-oss-120b}}                & MXFP4             & 4.25 & 117B    & 5.1B & 131k    \\
\href{https://huggingface.co/unsloth/Devstral-2-123B-Instruct-2512-GGUF}{\texttt{Devstral-2-123B-Instruct-2512}} & UD-Q8\_K\_XL    & 9.35 & 123B    & N/A & 256k     \\
\href{https://huggingface.co/unsloth/MiniMax-M2.5-GGUF}{\texttt{MiniMax-M2.5}}                & IQ5\_K          & 5.93 & 229B    & 10B & 196k     \\
%Qwen3.5-397B-A17B             & UD-Q8\_K\_XL & 397B     &  & 262k & \\
\href{https://huggingface.co/unsloth/DeepSeek-V3.2-GGUF}{\texttt{DeepSeek-V3.2}}               & Q4\_K\_M        & 4.83 & 685B    & 37B & 128k     \\
\href{https://huggingface.co/ubergarm/GLM-5.1-GGUF}{\texttt{GLM-5.1}}                & IQ4\_K          & 4.62      & 754B      & 40B & 202k  \\
\href{https://huggingface.co/AesSedai/Kimi-K2.5-GGUF}{\texttt{Kimi-K2.5}}                    & Q4\_X           & 4.55      & 1.1T      & 32B & 256k  \\
ChatGPT-5                     & N/A             & N/A       & N/A       & N/A & N/A   \\
Gemini-3                      & N/A             & N/A       & N/A       & N/A & N/A  \\
Claude Sonnet 4.6             & N/A             & N/A       & N/A       & N/A & N/A   \\ \bottomrule
\end{tabular}
\end{table*}

%Fuer das Quellenverzeichnis:
%https://huggingface.co/unsloth/Qwen3-Coder-Next-GGUF
%https://huggingface.co/unsloth/gpt-oss-120b-GGUF
%https://huggingface.co/unsloth/Devstral-2-123B-Instruct-2512-GGUF
%https://huggingface.co/unsloth/MiniMax-M2.5-GGUF
%https://huggingface.co/unsloth/DeepSeek-V3.2-GGUF
%https://huggingface.co/ubergarm/GLM-5.1-GGUF
%https://huggingface.co/AesSedai/Kimi-K2.5-GGUF

\subsection{Benchmark Suite \& Groundtruth Wrappers}

To assess the ability of large language models to generate correct simulation-interface layers we constructed a compact benchmark that covers the two target environments (WNTR and Modelica).

\paragraph{WNTR benchmark networks}
We used three water-distribu-tion examples that are shipped with the WNTR package.
For the purpose of this study we removed all control statements.
The resulting EPANAT \texttt{inp}-Files were then translated into our JSON schema.

%\begin{itemize}
%  \item \texttt{net1} – a minimal 2-node network (reservoir–junction–tank);
%  \item \texttt{net2} – a medium-size network with 13 junctions, two pumps and one valve;
%  \item \texttt{net3} – a larger branched network comprising 27 junctions, three pumps and two valves.
%\end{itemize}

\paragraph{Modelica benchmark modules}
For Modelica, we employed the four modules created by the HAI-CPPS benchmark \cite{Moddemann.2025}.
Each module originally contains components to simulate faults (e.g., valve leakages or a noisy pump controls) which were removed to simplify the problem. State-graphs for controlling the components were replaced by minimal control structure built from standard Modelica blocks.
The network structure was then converted to the JSON schema.

%\begin{enumerate}
%  \item \texttt{PumpSystem}
%  \item \texttt{ValveNetwork}
%  \item \texttt{MultiPumpPlant}
%  \item \texttt{FullPlant}
%\end{enumerate}

\paragraph{Ground-truth wrappers}
For each relevant component type in the neutral graph, we manually created a ground-truth simulation-interface layer that instantiates the corresponding object in the target framework and sets all of its parameters.
A second routine assembles these component functions, constructing a complete simulation model.
Consequently the two scripts realize the Ground-Truth (GT) mappings
$\mathcal{W}$ and $\mathcal{M}$ as defined in \cref{sec:method}.
The wrappers were deliberately kept minimal.

\subsection{Experiments}
\label{ref:experiments}

The generation task that we pose to the language models can be broken down into three implicit sub-tasks:
\begin{enumerate}
    \item \textbf{Unit conversion.} Numerical values in the neutral JSON graph (see \cref{sec:specification}) are expressed in generic SI units, whereas the target APIs (WNTR or the Modelica Standard Library) expect their own canonical units. The model must therefore recognize the required conversion and emit the correctly scaled literals. 
    \item \textbf{Simulation Interface Layer synthesis.} A complete, runnable module has to be produced, that creates the appropriate objects, connects them according to the graph topology and respects the idioms of the target language (naming, imports, formatting). 
    \item \textbf{Component-to-class mapping.} Each vertex type (pump, valve, tank, …) must be mapped to the corresponding class or block in the target library and all mandatory arguments have to be supplied. 
\end{enumerate}

The WNTR benchmarks feature larger-scale water distribution networks (9–92 junctions) with time-varying demand patterns, but employ simpler component types (pipes, pumps, tanks). 
In contrast, the Modelica benchmarks represent smaller industrial systems with more complex behaviors, including level control logic.

We evaluate a representative set of state-of-the-art large language models that differ in size, training focus (code-oriented vs. general-purpose) and maximum context length. 
The models used are Qwen3-Coder-Next \cite{Cao.2026}, gpt-oss-120b \cite{OpenAI.2025}, Devstral 2 \cite{Rastogi.2025}, Minimax M2.5 \cite{MiniMax.2026}, DeepSeek V3.2 \cite{DeepSeekAI.2025}, GLM 5.1 \cite{GLM5Team.2026}, Kimi K2.5 \cite{KimiTeam.2026}, ChatGPT-5 \cite{chatgpt5.2025}, Gemini-3 \cite{Pichai.2025} and Claude Sonnet 4.6 \cite{Anthropic.2026}. They were chosen due to being up-to-date and widely used. Additional details are listed in \cref{tab:models}. 
Furthermore, the larger models (Qwen3-Coder-Next, Devstral-2, Kimi-K2.5) are able to accommodate the full-source prompt (i.e., the complete code base).

For each model we run the generation pipeline with the five knowledge-integration strategies (Prompts 1-5) described in \cref{sec:prompts}, discarding a configuration if the model’s context window is insufficient to hold the prompt.
Due to computational restraints, we evaluate OpenCode with two models only, the higher performing model Kimi-K2.5 and the lighter, but lower performing model Qwen3-Coder-Next.
Small errors like missing parenthesis are corrected manually. 
Further, a single entry point~
\begin{figure}[h]
    \centering
    \begin{lstlisting}[breaklines]
    def main(network_dict: dict) -> Union[wntr.network.WaterNetworkModel, str]:
        ...
    \end{lstlisting}
\end{figure}

\noindent is added manually, which receives the neutral graph (as a Python dictionary) and returns either a WNTR network object or a Modelica file string.  
The resulting code artifacts are assessed with the suite of metrics defined in \cref{sec:metrics}.
To address the stochastic variability inherent in LLMs, each experimental configuration was executed three times.

\subsection{Experimental LLM Inference Setup}
To ensure full control over experimental parameters, all open-source models were hosted and executed on local hardware. 
This constraint introduces the significant challenge of fitting large models into the memory capacity of available hardware.

Modern LLMs are typically distributed in 16-bit floating-point precision (FP16) or 16 bits per weight (BPW), though exceptions exist.
For instance, Kimi-K2.5 is natively quantized to INT4 (4 BPW). 
The baseline memory requirement for a model is roughly the product of its parameter count and BPW, excluding the additional memory required for the Key-Value cache and inference overhead. 
For example, a 754-billion-parameter model like GLM-5.1 requires approximately 1.5 TB of memory in FP16 ($754\text{B} \times 16\text{ BPW}$), while a 1-trillion-parameter model at 4 BPW requires 512 GB. 
These demands exceed the available VRAM.

Quantization addresses this memory limitation by reducing the precision of the model weights, typically from 16 BPW to 8 or 4 BPW.
This reduction decreases both the memory footprint and memory bandwidth requirements, enabling the deployment of models on hardware that would otherwise be incapable of loading them \cite{Ashkboos.2024}.

The advanced quantization formats utilized in our experiments, namely \texttt{Q4\_K\_M} and \texttt{Q8\_K\_XL}, employ mixed-precision strategies.
In these schemes, critical network layers are preserved at higher bit-widths, while less sensitive tensors are more aggressively quantized.

To execute the quantized models efficiently across this heterogeneous architecture, we utilized llama.cpp \cite{ggmlorg.28.07.2026}, a highly optimized C/C++ inference engine. 
llama.cpp allows for dynamic distribution of the model workload. 
Specifically, for Mixture-of-Experts (MoE) architectures, the engine enables offloading the expert layers into the system's main memory (RAM), while keeping the active attention and dense layers on the GPU VRAM.
This hybrid offloading approach is critical for facilitating the execution of massive MoE models that exceed the VRAM ceiling, bridging the gap between computational speed and memory capacity.

\section{Experimental Results}
\label{sec:results}

In this section, we present the experimental results following the methodology outlined in \cref{sec:method}. 
We begin by analyzing the correlation between our evaluation metrics . 
Next, we compare the performance of the models listed in \cref{tab:models} across the five prompting configurations (1–5) in \crefrange{sec:model_performance}{sec:context}. 
Finally, we evaluate the use of coding agents in \cref{sec:opencode}.

\subsection{Correlation of Quality Metrics}

\begin{figure}[tb]
    \centering
    \includegraphics[width=0.99\linewidth]{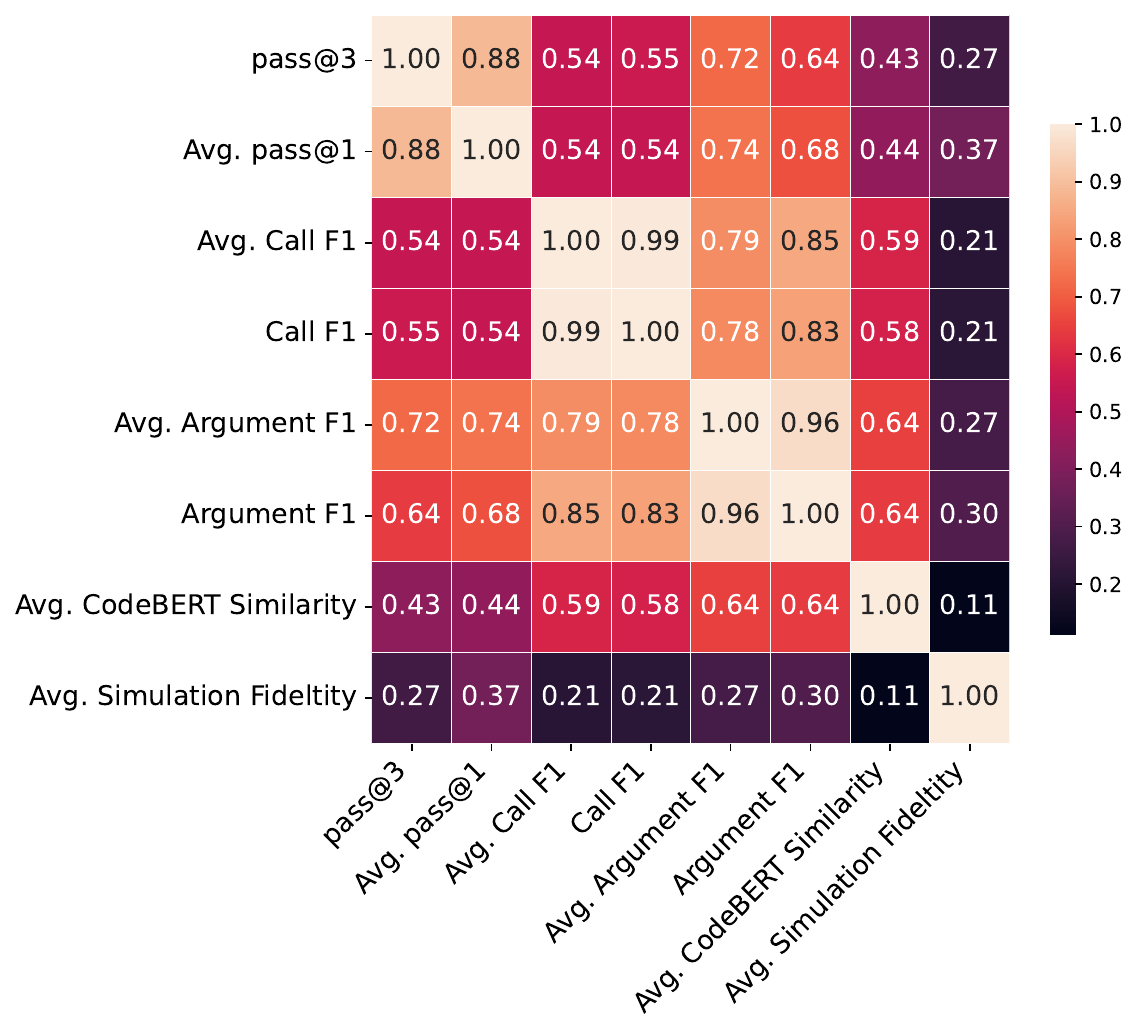}
    \caption{Pearson correlation matrix (absolute values) of evaluation metrics. Darker cells indicate weaker absolute (linear) correlation.}
    \label{fig:results_correlation}
\end{figure}

To identify redundant or complementary metrics, we computed Pearson correlation coefficients across all evaluation dimensions and display the results in \cref{fig:results_correlation}, as done by, e.g., \citet{Pereira.2018}.
The analysis reveals that global (averaged) and local variants of Call F1 and Argument F1 scores are nearly identical, indicating that per-wrapper and aggregate metrics capture mostly the same underlying trends. 
CodeBERT similarity shows negligible correlation with other metrics, suggesting it measures orthogonal aspects of code quality (e.g., stylistic rather than functional correctness).
Simulation fidelity is not strongly correlated with any other metrics and therefore an important metric assessing the whole code generation quality.

Given these observations, we focus on pass@1, pass@3, Call F1, Argument F1, and Avg. Simulation Fidelity, which together cover complementary aspects of code generation and simulation performance while avoiding unnecessary duplication in the evaluation.

\begin{figure*}[tb]
    \centering
    \includegraphics[width=0.99\linewidth]{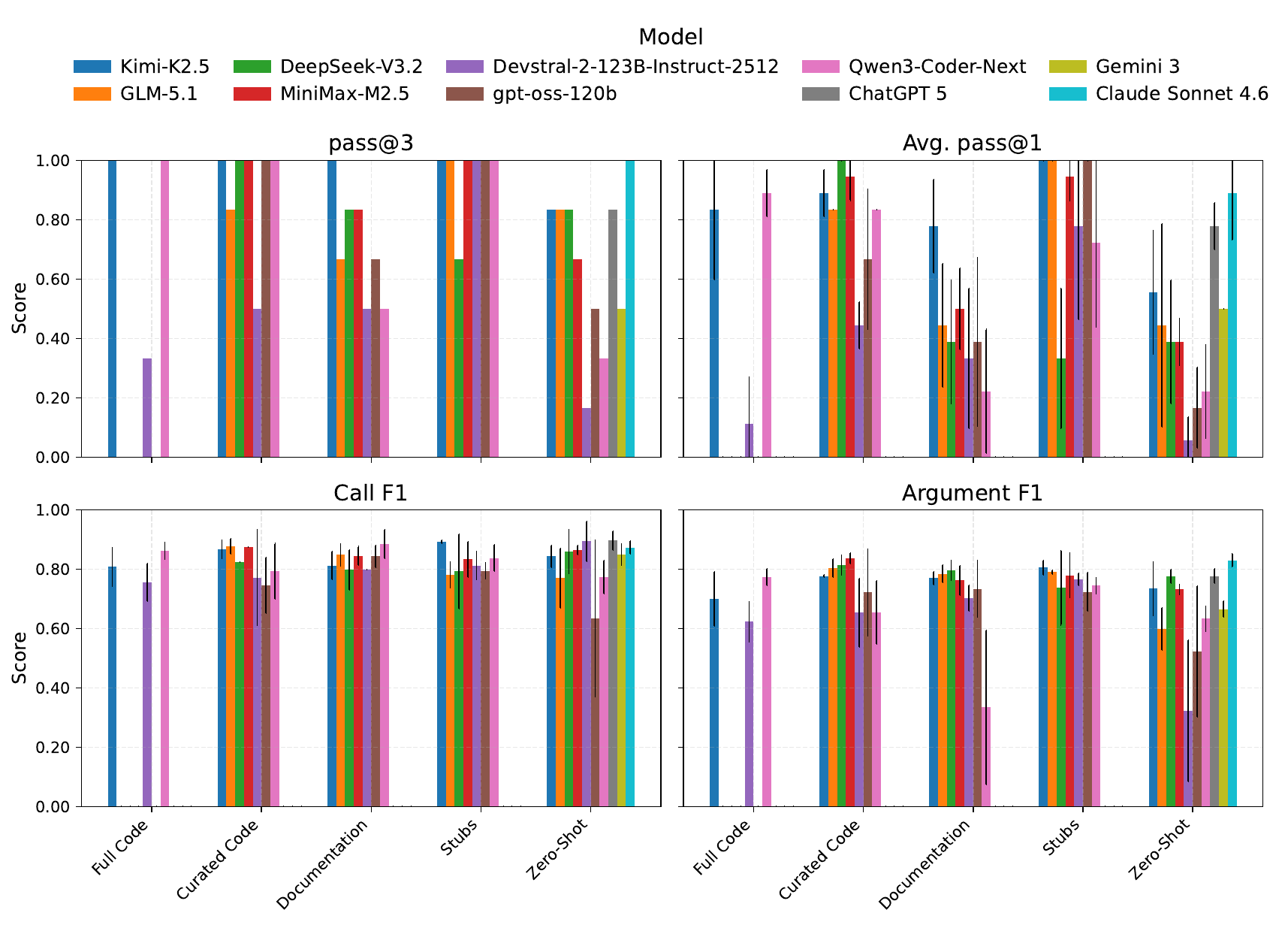}
    \caption{Performance of LLMs on WNTR wrapper generation. Each model is run for prompt configurations 1 (zero-shot) to 5 (code base). Missing values indicate configurations exceeding context limits or producing non-executable code.}
    \label{fig:results_wntr}
\end{figure*}

\begin{figure*}[tb]
    \centering
    \includegraphics[width=0.99\linewidth]{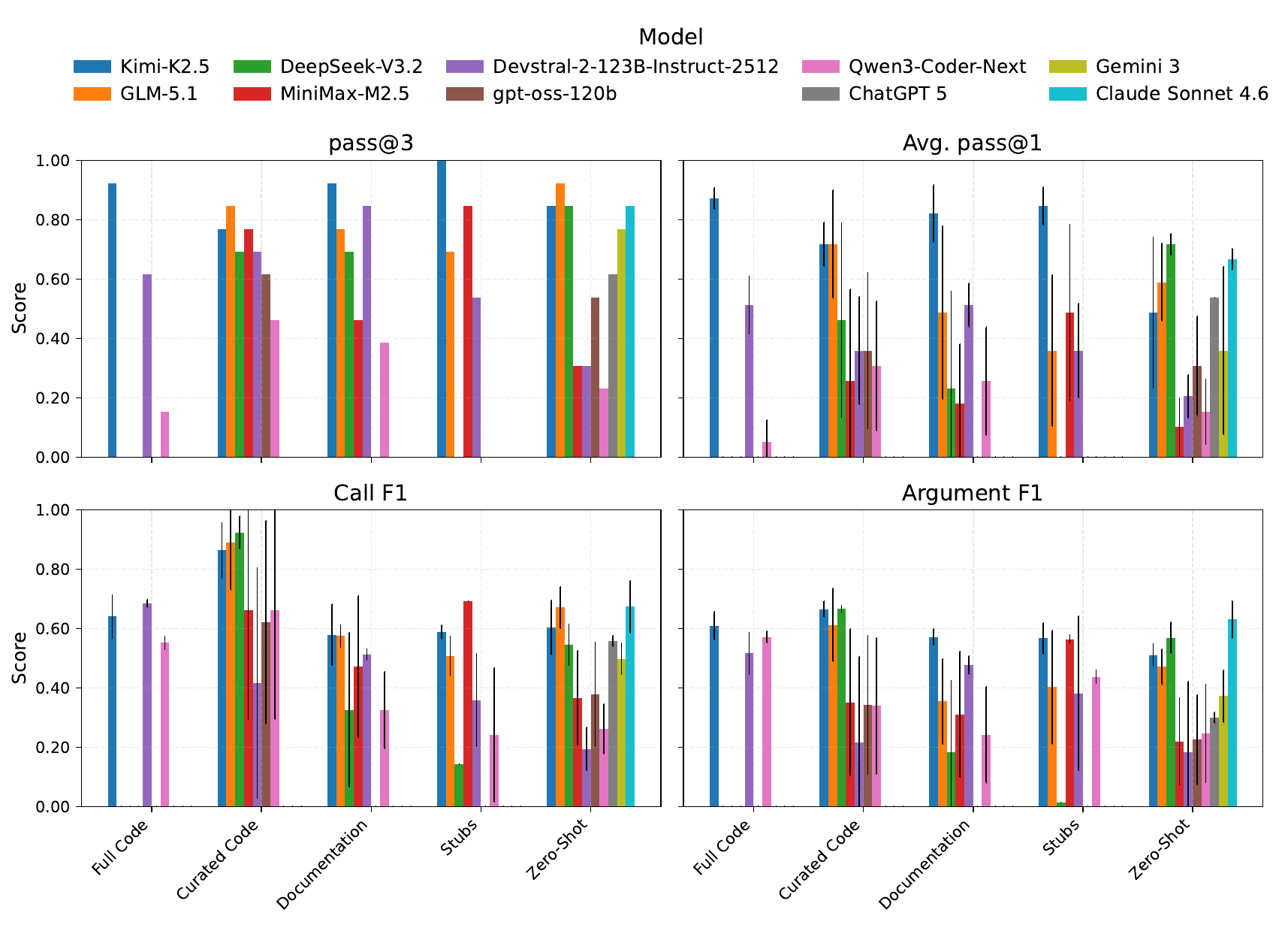}
    \caption{Performance of LLMs on Modelica wrapper generation. Each model is run for prompt configurations 1 (zero-shot) to 5 (code base). The frequent generation failures highlight the challenges of synthesizing code for strongly-typed, equation-based modeling languages that differ fundamentally from more common programming languages like Python.}
    \label{fig:results_modelica}
\end{figure*}

\subsection{Model Performance}
\label{sec:model_performance}
When comparing models across the prompting strategies 1-5 and target environments WNTR and Modelica (\cref{fig:results_wntr} and \cref{fig:results_modelica}), Kimi-K2.5 stands out with high scores.
For the Modelica environment, it achieves pass@1 up to 0.87, pass@3 up to 1, with high F1 scores reaching around 0.86. 
On WNTR, it frequently reaches perfect pass@1 and pass@3 scores of 1.0, with very high F1 scores between 0.85-0.89. 
This model shows consistency across different prompt types and is least sensitive to prompt variation, excelling in both syntactic correctness and API conformity. 
Kimi-K2.5 stands out as the only model that achieves near-ceiling performance on WNTR while maintaining strong, though not perfect, results on Modelica.

GLM-5.1 ranks as the second strongest model, showing particular robustness across both frameworks. 
In Modelica, it reaches pass@1 up to 0.72 and pass@3 up to 0.92, with strong F1 scores peaking at approximately 0.89. 
For WNTR, it achieves perfect pass@1 scores and maintains very stable high performance across different prompts. While less extreme than Kimi-K2.5, GLM-5.1 demonstrates stability across prompt types and strong generalization capabilities.
Among the weaker performers, Qwen3-Coder-Next shows significant domain sensitivity. 
It struggles with Modelica, achieving low pass@1 scores between 0.07-0.25, though its pass@3 can reach approximately 0.69, indicating some search ability despite weak precision. 
On WNTR, it performs much better with pass@1 up to 0.89, though still below the top models. 
This pattern suggests Qwen3-Coder-Next has particular difficulty with Modelica's constrained syntax while performing better in WNTR's Python environment.
gpt-oss-120b demonstrates moderate performance across both domains. 
For Modelica, it achieves pass@1 around 0.3 and pass@3 up to 0.61, with weak all-pass rates and consistency. 

Additionally, weaker models like Qwen3-Coder-Next, MiniMax-M2.5, and Devstral-2-123B show large gaps between pass@1 and pass@3, suggesting they find correct code only through multiple attempts.
In contrast, top models like Kimi-K2.5 and GLM-5.1 show much smaller gaps, indicating they get the code right with higher consistency.

Despite conducting only three runs per configuration, the observed standard deviations reveal significant variability in model performance across repetitions.
The standard deviations in our results reveal notable variability in model performance across multiple runs, particularly for less capable models. 
For Modelica, we observe high STDs (e.g., up to 0.326 for pass@1 in DeepSeek-V3.2 with code context), indicating inconsistency in generating valid simulation code. 
In contrast, top-performing models like Kimi-K2.5 exhibit lower STDs (often <0.1), demonstrating more stable performance. 
For WNTR, the variability is generally lower, suggesting that Python-based code generation is more deterministic for most models. 
This pattern underscores that while leading models achieve consistent results, others struggle with reliability, especially in complex domains like Modelica.

\begin{figure*}[tb]
    \centering
    \includegraphics[width=0.99\linewidth]{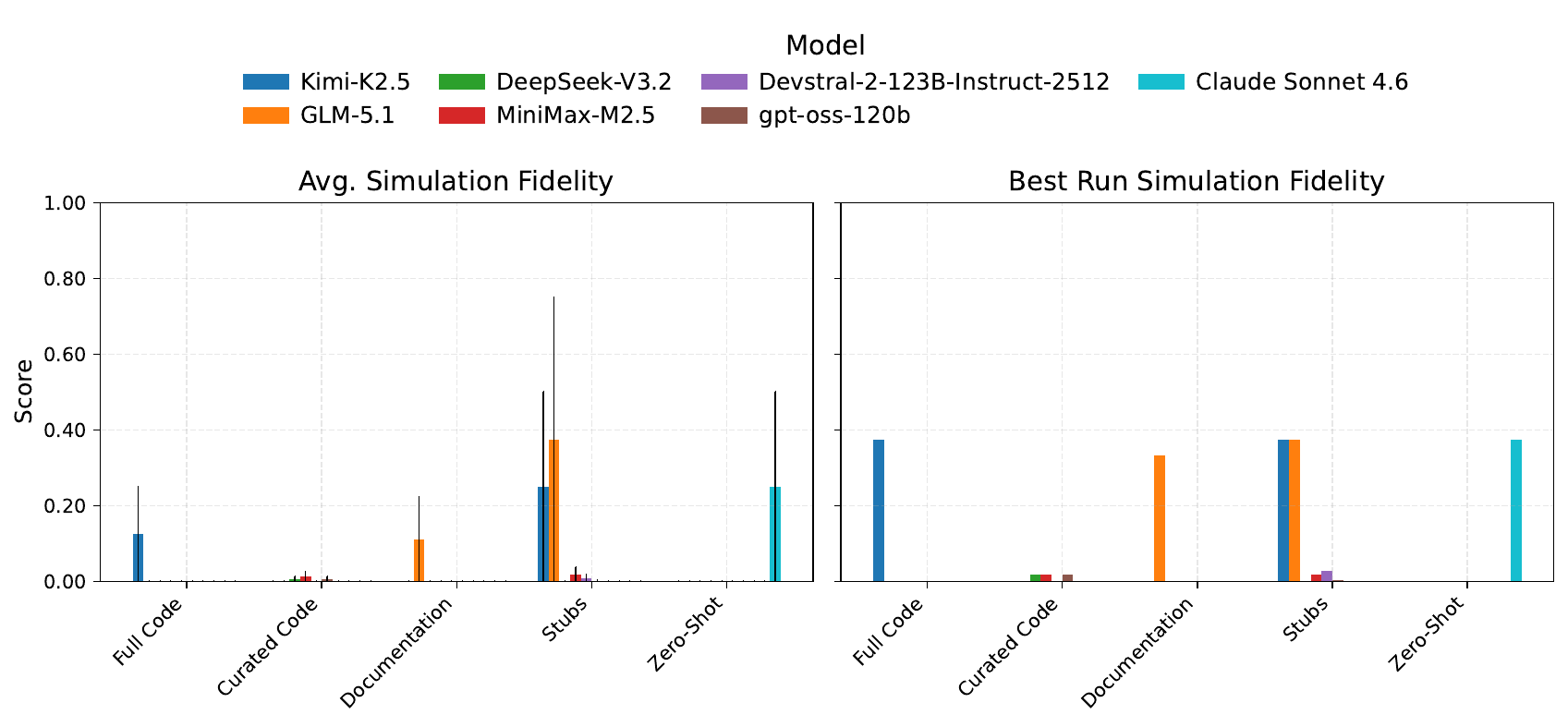}
    \caption{WNTR simulation fidelity results: (left) average across three runs, (right) best of three runs. The consistently low fidelity indicates LLMs struggle to generate physically accurate simulation models.}
    \label{fig:results_wntr_sim}
\end{figure*}

\subsection{Domain Comparison: Impact on the target domain}

We compare the impact w.r.t. target environments using the models of \cref{tab:models} using prompting strategies 1 to 5.
The empirical evaluation in \cref{sec:model_performance} reveals a heterogeneous performance landscape: the same language model can excel on the Python-based WNTR benchmark while collapsing to zero on the Modelica DSL, and vice-versa.  
The most affected metric is simulation fidelity, which is unavailable for the majority of Modelica runs because the generated files do not compile or the wrapper crashes before the simulation can be launched.

Pass rates are consistently higher for WNTR as seen in \cref{fig:results_wntr}, with pass@1 means exceeding Modelica (\cref{fig:results_modelica} by up to 20\%, and pass@3 and pass@all showing even larger differences. 
F1 scores for both call and argument correctness are also consistently higher on WNTR, while variance is often lower or comparable. 
Notably, weaker models like Devstral-2-123B-Instruct-2512 that struggle with Modelica (achieving pass@1 below 0.3-0.5) often reach 0.7-1.0 pass@1 on WNTR with stronger prompts.
While no Modelica configuration produced directly simulatable models, some WNTR configurations achieved partial simulation success. 
However, the results were inconsistent because the same models sometimes failed completely, ending in high variance across runs.
The results clearly show WNTR is systematically easier for LLMs than Modelica across all metrics. 

\subsection{Context: Impact of the prompting strategies 1-5}
\label{sec:context}

Evaluating the influence of the prompting strategies shows that when given the exact functions to use as context (curated prompt), all models have the highest pass@1 scores and best F1 metrics across both domains.
This advantage is particularly evident in Modelica, where GLM-5.1 achieves pass@1 of approximately 0.49 with curated prompts compared to 0.36 with documentation and even lower with other context approaches. 
The effect is even more pronounced for WNTR, where stronger models like GLM-5.1 and Kimi-K2.5 reach near-optimal performance under curated prompts.
Full code context does not improve upon stub-based approaches, suggesting that providing complete codebases may introduce noise rather than information. 
Zero-shot performance shows variation across models, with many collapsing to a 0.1-0.3 range while Claude Sonnet 4.6 achieves surprisingly high values.

\subsection{OpenCode}
\label{sec:opencode}

We conducted experiences using Coding Agents as described in \cref{sec:opencode_experiment}.
We compare the code created with OpenCode with and without validation function with the results from prompt configuration 5 (full code).
The results can be seen in \cref{tab:modelica_opencode} for Modelica and in \cref{tab:wntr_opencode} for WNTR.
While OpenCode alone does not outperform direct LLM generation with codebase context, the addition of the validation function yields consistently superior results. 
OpenCode with validation achieves perfect pass@3 rates (1.0 and 0.923) and substantially improves pass@1 scores, reaching 0.923 for Kimi-K2.5 and 0.846 for Qwen3-Coder-Next. 
This configuration produces both compilable and simulatable Modelica code, though simulation fidelity remains at zero across all attempts.

The validation function appears crucial for success, enabling the agent to correct syntactic and API-related errors through iterative refinement. However, the persistent zero simulation fidelity indicates that while the agent can fix surface-level code issues, it cannot resolve deeper semantic problems related to physical parameter mappings or equation-based logic.

\begin{table*}[tb]
\small
\centering
\caption{Comparison of Modelica code generation performance between direct LLM generation, OpenCode agent, and OpenCode with validation function. While validation improves syntactic correctness and API conformity, simulation fidelity remains zero across all configurations.}
\label{tab:modelica_opencode}
\begin{tabular}{@{}rccccc@{}}
\toprule
\multicolumn{1}{c}{}                                                                           & \multicolumn{1}{c}{pass@3} & \multicolumn{1}{c}{pass@1} & \multicolumn{1}{c}{Call F1} & \multicolumn{1}{c}{Arg. F1} & \multicolumn{1}{c}{\begin{tabular}[c]{@{}c@{}}Avg. \\ Simulation \\ Fidelity\end{tabular}} \\ \midrule
Kimi-K2.5                                                                                      & 0.923                      & $0.872 \pm 0.036$          & $\mathbf{0.641 \pm 0.072}$  & $0.610 \pm 0.047$           & $0.0 \pm 0.0$                                                                              \\
OpenCode Kimi-K2.5                                                                             & 0.846                      & $0.692 \pm 0.063$          & $0.528 \pm 0.091$           & $0.587 \pm 0.032$           & $0.0 \pm 0.0$                                                                              \\
\begin{tabular}[c]{@{}r@{}}OpenCode Kimi-K2.5\\ (with validation function)\end{tabular}        & \textbf{1.000}             & $\mathbf{0.923 \pm 0.063}$ & $0.606 \pm 0.045$           & $\mathbf{0.639 \pm 0.016}$  & $0.0 \pm 0.0$                                                                              \\
\multicolumn{1}{l}{}                                                                           &                            &                            &                             &                             &                                                                                            \\
Qwen3-Coder-Next                                                                               & 0.692                      & $0.462 \pm 0.218$          & $0.568 \pm 0.011$           & $0.488 \pm 0.040$           & $0.0 \pm 0.0$                                                                              \\
OpenCode Qwen3-Coder-Next                                                                      & 0.308                      & $0.256 \pm 0.036$          & $0.269 \pm 0.039$           & $0.329 \pm 0.049$           & $0.0 \pm 0.0$                                                                              \\
\begin{tabular}[c]{@{}r@{}}OpenCode Qwen3-Coder-Next\\ (with validation function)\end{tabular} & \textbf{0.923}             & $\mathbf{0.846 \pm 0.063}$ & $\mathbf{0.615 \pm 0.0}$    & $\mathbf{0.588 \pm 0.029}$  & $0.0 \pm 0.0$                                                                              \\ \bottomrule
\end{tabular}
\end{table*}

\begin{table*}[tb]
\centering
\small
\caption{Comparison of WNTR code generation performance showing similar trends, where OpenCode with validation achieves the highest syntactic correctness but still struggles with simulation fidelity.}
\label{tab:wntr_opencode}
\begin{tabular}{@{}rccccc@{}}
\toprule
 & pass@3 & pass@1                     & Call F1                    & Arg. F1                    & \multicolumn{1}{c}{\begin{tabular}[c]{@{}c@{}}Avg.\\ Simulation\\ Fidelity\end{tabular}} \\ \midrule
Kimi-K2.5 & 1.000  & $0.833 \pm 0.236$          & $0.808 \pm 0.065$          & $0.700 \pm 0.091$          & $0.125 \pm 0.125$ \\
OpenCode Kimi-K2.5 & 1.000  & $0.889 \pm 0.157$          & $0.834 \pm 0.069$          & $0.755 \pm 0.036$          & $0.131 \pm 0.131$ \\
\begin{tabular}[c]{@{}r@{}}OpenCode Kimi-K2.5 \\ (with validation function)\end{tabular}       & 1.000  & $\mathbf{0.944 \pm 0.079}$ & $\mathbf{0.835 \pm 0.076}$ & $\mathbf{0.815 \pm 0.026}$ & $\mathbf{0.375 \pm 0.375}$                                                               \\
 & & & & &                                                                                          \\
Qwen3-Coder-Next & 1.000  & $0.889 \pm 0.079$          & $0.862 \pm 0.028$          & $0.773 \pm 0.028$          & $0.000 \pm 0.000$ \\
OpenCode Qwen3-Coder-Next & 1.000  & $0.889 \pm 0.157$          & $0.856 \pm 0.087$          & $0.754 \pm 0.062$          & $0.015 \pm 0.015$ \\
\begin{tabular}[c]{@{}r@{}}OpenCode Qwen3-Coder-Next\\ (with validation function)\end{tabular} & 1.000  & $\mathbf{0.944 \pm 0.079}$ & $\mathbf{0.900 \pm 0.058}$ & $\mathbf{0.805 \pm 0.016}$ & $\mathbf{0.063 \pm 0.063}$ \\ \bottomrule
\end{tabular}
\end{table*}

\subsection{Ablation}
To assess the models' ability to handle end-to-end simulation code generation, we conducted an ablation experiment where we tasked the LLMs with directly producing complete Modelica files for the first benchmark network, rather than generating modular wrapper functions. 
Despite this simplified task formulation, none of the evaluated models succeeded in producing syntactically valid or translatable Modelica code.
Such an experiment was not conducted with WNTR, as it is implemented in Python and thus does not require an additional translation step.

\section{Discussion}
\label{sec:discussion}

The results suggest that domain familiarity outweighs raw model size.
The 80B-parameter, code-specialized model Qwen3-Coder-Next outperforms big models on the Python-based WNTR benchmark, yet it fails almost completely on Modelica.
Conversely, big models such as Kimi-K2.5 achieve good scores on both benchmarks.
This suggests that exposure to the target language during training (Python vs. Modelica) is the dominant factor, not the number of parameters.  

Another observation is that the size of the context window (128k-262k tokens for all models) does not correlate with generation quality.
Within the window, relevance is far more important than volume.
Curated prompts that contain a minimal set of API snippets needed for the task consistently outperform prompts that use the entire code base or documentation.
The finding follows other work, where LLMs struggle to extract useful information from very long contexts \cite{Liu.2024}.
This could also explain why zero-shot prompting sometimes matches or even exceeds prompting the documentation. 
The model relies on its internal knowledge rather than on (noise or irrelevant) external material.  

The low Call F1 scores observed for Modelica could be attributed to the ambiguity of graph-to-simulation mapping.
The graph only labels a component as \texttt{valve} or \texttt{pump} but the Modelica Standard Library contains multiple classes for each of these components.
When the language model selects the wrong class or function the generated call is counted as a miss, which drags the Call F1 down.
Argument F1 follows the same pattern because arguments are only evaluated when the surrounding call is present.
A missing or incorrect call automatically also penalizes the argument score.

Using agents (OpenCode) together with a validation function that attempts to compile the generated Modelica file improves syntactic correctness and raises pass@1/3, Call F1 and Argument F1 scores.
The agent can repair missing parentheses, wrong imports, or mismatched argument names, but it does not resolve deeper semantic errors such as incorrect component selection or unit conversion.
Consequently simulation fidelity remains at zero for all configurations.  

While our evaluation framework provides systematic measurements of LLM-generated simulation code, several methodological limitations warrant discussion.
The automatic evaluation pipeline introduces problems and therefore noise in the results. For example, despite an explicit instruction in the prompt to avoid class-based wrappers, many models still emit methods inside a class.
Our parser does not recognize these as valid calls, so they are not counted and thus the metrics are lowered.  
Additionally, not all evaluation metrics proved equally informative. 
The CodeBERT-based similarity measurement, intended to assess semantic alignment between synthesized and reference implementations, yielded limited discriminative value. 

The observed difficulties in generating physically accurate simulation models highlight a fundamental limitation of LLMs, which is their inability to inherently understand or reason about physical laws and algorithmic constraints.
While LLMs excel at pattern recognition and syntactic code generation, they lack the capacity to internalize domain-specific knowledge such as fluid dynamics principles or control system algorithms. 
This aligns with findings from \citet{Jin.2026}, which demonstrate that LLMs frequently misclassify correct implementations when faced with domain-specific requirements.

\paragraph{Practical Advice for Simulation Experts}
From the results, we can make the following recommendations when trying to build model-based systems using LLMs:
\begin{itemize}
    \item Use Kimi-K2.5, as it demonstrated the most consistent performance across both WNTR and Modelica benchmarks.
    \item Carefully narrow down the context by manually selecting only the most relevant components and functions to avoid overwhelming the model with irrelevant information.
    \item Preselecting relevant functions and code snippets yields better outcomes than providing full codebases or extensive documentation.
    \item Always expect to perform manual validation of generated code, particularly for complex domains like Modelica where semantic correctness is critical.
    \item When possible, employ coding agents with validation functions to iteratively refine and verify the generated code.
\end{itemize}

\section{Conclusion}
\label{sec:conclusion}

This study shows both the promise and current limitations of LLM-driven simulation code generation for model-based systems engineering.
While state-of-the-art models can reliably produce syntactically correct wrappers for the Python-based tool WNTR, generating executable Modelica code remains challenging due to the domain-specific nature of equation-based modeling languages. 
Our findings reveal that performance depends more on domain familiarity than model size, with code-specialized models outperforming larger general-purpose models in their trained domains.

While fully automated simulation code generation remains an aspirational goal, our results show that LLMs can already significantly accelerate model-based engineering workflows when used with caution. 
Even though simulations are runnable, they are not necessarily correct.
Currently, the key to practical adoption lies in understanding the current limitations and implementing appropriate safeguards, rather than expecting perfect end-to-end automation.

\section*{Acknowledgments}
We thank Jan Kels for valuable discussions on LLM-based code generation techniques and code generation metrics.

\section*{Declaration of generative AI and AI-assisted technologies in the manuscript preparation process}
During the preparation of this work the author(s) used gpt-oss-120B and Mistral Large 3 for assistance with language refinement and technical formulation. After using this tool/service, the authors reviewed and edited the content as needed and takes full responsibility for the content of the published article.

% Reducing font size (to 9pt) for References & Author Biagraphies
\footnotesize

% Please don't exchange the bibliographystyle style
\bibliographystyle{elsarticle/elsarticle-num-names}

% AUTHOR: Include your bib file here
\bibliography{literature}

\end{document}